\documentclass[letterpaper]{article}

\usepackage{times}
\usepackage{helvet}
\usepackage{courier}
\usepackage{arxiv}

\usepackage[utf8]{inputenc} 
\usepackage[T1]{fontenc}    
\usepackage{hyperref}       
\usepackage{url}            
\usepackage{booktabs}       
\usepackage{amsfonts}       
\usepackage{amsmath}
\usepackage{subfig}
\usepackage{nicefrac}       
\usepackage{microtype}      
\usepackage{graphicx}
\usepackage{natbib}
\usepackage{doi}

\title{Representation of the structure of graphs by sequences of instructions}

\author{ \href{https://orcid.org/0000-0001-8231-5687}{\includegraphics[scale=0.06]{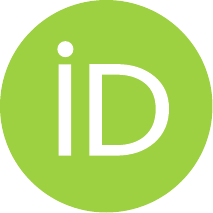}\hspace{1mm}Ezequiel L\'opez-Rubio}\thanks{Corresponding author. ITIS Software. Universidad de M\'alaga. C/ Arquitecto Francisco Peñalosa 18, 29010, Málaga, Spain} \\
	Department of Computer Languages and Computer Science\\
    University of M\'alaga\\
    Bulevar Louis Pasteur, 35\\
    29071 M\'alaga, Spain \\
	\texttt{ezeqlr@lcc.uma.es} \\
}

\renewcommand{\shorttitle}{Representation of graphs by sequences of instructions}

\hypersetup{
pdftitle={Representation of the structure of graphs by sequences of instructions},
pdfsubject={cs.AI, cs.CL},
pdfauthor={Ezequiel L\'opez-Rubio},
pdfkeywords={graph representation,  adjacency matrix, instruction sequences, deep learning, language models, structural
patterns}
}

\begin{document}
\maketitle

\begin{abstract}
The representation of graphs is commonly based on the adjacency matrix
concept. This formulation is the foundation of most algebraic and
computational approaches to graph processing. The advent of deep learning
language models offers a wide range of powerful computational models
that are specialized in the processing of text. However, current procedures
to represent graphs are not amenable to processing by these models.
In this work, a new method to represent graphs is proposed. It represents
the adjacency matrix of a graph by a string of simple instructions.
The instructions build the adjacency matrix step by step. The transformation
is reversible, i.e., given a graph the string can be produced and vice
versa. The proposed representation is compact, and it maintains the
local structural patterns of the graph. Therefore, it is envisaged
that it could be useful to boost the processing of graphs by deep
learning models. A tentative computational experiment is reported,
demonstrating improved classification performance and faster
computation times with the proposed representation.
\end{abstract}


\section{Introduction}

Graphs provide a flexible abstraction for relational data in domains such as social networks, molecules, knowledge graphs, recommendation, and databases \citep{zhou2020gnn,khoshraftar2022survey,ju2023deep}. The standard approach to graph processing with deep learning models is to learn a suitable representation of them. The goal of graph representation learning is to map nodes, edges, subgraphs, or whole graphs into low-dimensional vectors that preserve structural properties and attributes, enabling downstream tasks such as node classification, link prediction, graph classification, and anomaly detection \citep{khoshraftar2022survey,ju2023deep}.

Early work on graph representation focused on \emph{shallow} embedding methods that learn a lookup table of node embeddings optimized for proximity in the original graph \citep{khoshraftar2022survey}. Random-walk-based methods such as DeepWalk and node2vec treat truncated random walks as sentences and apply word embedding techniques to enforce that co-visited nodes obtain similar vectors \citep{perozzi2014deepwalk,grover2016node2vec}. Matrix factorization approaches, including Laplacian eigenmaps and variants based on factorizing pointwise mutual information matrices, can be interpreted as implicitly optimizing similar proximity objectives \citep{belkin2003laplacian,ou2016asymmetric}. These techniques are scalable and effective but decouple representation learning from node features and struggle to generalize to unseen nodes or dynamic graphs \citep{khoshraftar2022survey,ju2023deep}.

Extensions of shallow embeddings incorporate side information and edge types, e.g., for heterogeneous and knowledge graphs. Knowledge graph embedding methods such as TransE, DistMult, and RotatE embed entities and relations into continuous spaces and define scoring functions for triplets \citep{wang2017knowledge}. While powerful for link prediction, these models typically ignore higher-order structure and are limited in expressivity compared to modern deep architectures \citep{wang2017knowledge,ju2023deep}.

Deep graph representation learning is now dominated by Graph Neural Networks (GNNs), which implement message passing over the graph structure \citep{zhou2020gnn,wu2020comprehensive}. In the standard message-passing framework, each node iteratively aggregates information from its neighbors and updates its hidden state using a permutation-invariant function, yielding embeddings that combine local structure and node features \citep{gilmer2017mpnn,zhou2020gnn}. Popular instances include Graph Convolutional Networks (GCN), GraphSAGE, and Graph Attention Networks (GAT), which differ mainly in their neighborhood aggregation and normalization schemes \citep{kipf2017semi,hamilton2017inductive,velivckovic2018graph}.

GNNs can be categorized by their architectural principles \citep{zhou2020gnn,ju2023deep}. Spectral GNNs define convolutions via the graph Laplacian eigenbasis, while spatial GNNs perform aggregation directly in the vertex domain using learned filters \citep{bruna2014spectral,kipf2017semi}. Recurrent and attention-based variants replace simple aggregators with recurrent units or attention mechanisms to capture more expressive interactions \citep{gilmer2017mpnn,velivckovic2018graph}. Recent work connects GNN expressivity to the Weisfeiler–Lehman (WL) test, leading to architectures such as Graph Isomorphism Networks (GIN) that match the discriminative power of the 1-WL test under certain conditions \citep{xu2019gin}.

In this work, a completely different approach is taken. Rather than learning a representation, a new, fixed way to represent graphs is proposed. The approach is based on representing the adjacency matrix of the graph by a string of simple instructions. Our overall goal is to design a graph representation methodology that is amenable to processing by deep language models.

The structure of this paper is as follows. Section \ref{sec:Methodology}
presents the graph representation methodology. After that, Section
\ref{sec:Computational-experiments} reports the results of an exploratory
computational experiment. Finally, Section \ref{sec:Conclusion} deals
with the conclusions.

\section{Methodology\label{sec:Methodology}}

In this section, the proposed methodology to represent the structure
of a graph is detailed. Let $G=\left(V,A\right)$ be a graph where
$V=\left\{ v_{1},...,v_{N}\right\} $ is the set of vertices, and
$A$ is the set of edges $\left(v_{i},v_{j}\right)$, with $v_{i},v_{j}\in V$.
It will be assumed that a complete order is defined on the $N$ elements
(vertices) in $V$, i.e., $v_{1}$ is the first vertex, up to the last
vertex $v_{N}$. Then, the adjacency matrix associated with $G$ will
be noted $M_{G}$, which has $N\times N$ elements:

\begin{equation}
M_{G}\left(i,j\right)=\begin{cases}
0 & \textrm{if }\left(v_{i},v_{j}\right)\in A\\
1 & \textrm{if }\left(v_{i},v_{j}\right)\notin A
\end{cases}
\end{equation}

The adjacency matrix $M_{G}$ is symmetric for undirected graphs,
while it may not be symmetric for directed graphs. It comprises the
structure of the graph $G$. A standard way to represent such a structure
by a string is to flatten the adjacency matrix into a binary string
row by row:

\begin{equation}
B_{G}=M_{G}\left(1,1\right)\;...\;M_{G}\left(1,N\right)\;M_{G}\left(2,1\right)\;...\;M_{G}\left(N,N\right)
\end{equation}
where $B_{G}\in\left\{ 0,1\right\} ^{*}$ is a string of $N^{2}$
symbols.

\subsection{String representation}

Next, an alternative method to represent the structure of $G$ is
given. A string $w\in\left\{ U,D,L,R,E\right\} ^{*}$ is associated
with $G$, where the five possible symbols are interpreted as instructions
that can be executed to build $M_{G}$ starting from the null matrix
of size $N\times N$. A pointer $\mathbf{p}=\left(p_{1},p_{2}\right)\in\left\{ 1,...,N\right\} \times\left\{ 1,...,N\right\} $
to an element of $M_{G}$ will be maintained, which starts at position
$\left(1,1\right)$. The string $w$ is executed symbol by symbol,
from left to right. The current and next values of the pointer are
noted $\mathbf{p}$ and $\mathbf{p}'$, while the current and next
values of the adjacency matrix are noted $M$ and $M'$, respectively.
The semantics (meaning) of the instructions are:
\begin{itemize}
\item $U$ moves the pointer up if possible:\\
\begin{equation}
\mathbf{p}'=\begin{cases}
\mathbf{p} & \textrm{if }p_{1}=1\\
\left(p_{1}-1,p_{2}\right) & \textrm{if }p_{1}>1
\end{cases}
\end{equation}
\item $D$ moves the pointer down if possible:\\
\begin{equation}
\mathbf{p}'=\begin{cases}
\mathbf{p} & \textrm{if }p_{1}=N\\
\left(p_{1}+1,p_{2}\right) & \textrm{if }p_{1}<N
\end{cases}
\end{equation}
\item $L$ moves the pointer left if possible:\\
\begin{equation}
\mathbf{p}'=\begin{cases}
\mathbf{p} & \textrm{if }p_{2}=1\\
\left(p_{1},p_{2}-1\right) & \textrm{if }p_{2}>1
\end{cases}
\end{equation}
\item $R$ moves the pointer right if possible:\\
\begin{equation}
\mathbf{p}'=\begin{cases}
\mathbf{p} & \textrm{if }p_{2}=N\\
\left(p_{1},p_{2}+1\right) & \textrm{if }p_{2}<N
\end{cases}
\end{equation}
\item $E$ inserts a new edge at the pointer position:\\
\begin{equation}
M'\left(i,j\right)=\begin{cases}
1 & \textrm{if }\left(i=p_{1}\right)\land\left(j=p_{2}\right)\\
M\left(i,j\right) & \textrm{if }\left(i\neq p_{1}\right)\lor\left(j\neq p_{2}\right)
\end{cases}
\end{equation}
\\
where the element $M'\left(j,i\right)$ is also set to 1 if $G$ is
defined as an undirected graph.
\end{itemize}
The execution of any string in $\left\{ U,D,L,R,E\right\} ^{*}$ produces
an adjacency matrix. In other words, the set of valid strings is the
regular language $\left\{ U,D,L,R,E\right\} ^{*}$. Conversely, given
an adjacency matrix, there are infinite strings $w\in\left\{ U,D,L,R,E\right\} ^{*}$
that can produce it. For example, one can add pointer movement instructions
that cancel each other. In order to have a canonical string to represent
each graph, an algorithm is provided. The algorithm accepts an adjacency
matrix $M$ as input, produces a string $w\in\left\{ U,D,L,R,E\right\} ^{*}$
as output, and keeps a local variable $\mathbf{p}$ that is a pointer
to an element of $M$.
\begin{enumerate}
\item Initialize the pointer $\mathbf{p}$ to $\left(1,1\right)$, i.e the
upper left corner of $M$.
\item Initialize $w$ to the empty string.
\item If $M$ is null then halt and return $w$. Otherwise, go to step 4.
\item Find the nonzero cell $\mathbf{q}\in\left\{ 1,...,N\right\} \times\left\{ 1,...,N\right\} $
of $M$ which is closest to $\mathbf{p}$ according to Manhattan distance.
If there is more than one closest nonzero cells, then choose the
cell for which the difference $q_{1}-p_{1}$ is smallest.
\item Append $U$ or $D$ instructions to $w$ as required to move the pointer
$\mathbf{p}$ to the same row as $\mathbf{q}$.
\item Append $L$ or $R$ instructions to $w$ as required to move the pointer
$\mathbf{p}$ to the same column as $\mathbf{q}$.
\item Append instruction $E$ to $w$.
\item Set the element of $M$ pointed by $\mathbf{p}$ to zero. If $G$
is undirected, then set its symmetric element to zero too.
\item Go to step 3.
\end{enumerate}
The above algorithm follows a greedy approach to find short substrings
of pointer movement instructions so that the overall length of the
canonical representation of the input graph is kept small. There is
no guarantee that this is the shortest possible string to represent
$G$, but it is guaranteed that for each graph $G$ there is a unique
canonical string that represents it, namely the output of the above
algorithm. Let us note $I_{G}\in\left\{ U,D,L,R,E\right\} ^{*}$ the
canonical string for graph $G$.

\subsection{Representation length}

The length of the canonical string is upper bounded by $2N^{2}-1$.
The worst case is an adjacency matrix of a complete directed graph,
i.e. all elements of the matrix are one. In this case, each element
requires one pointer movement instruction plus one $E$ instruction
except the first element $\left(1,1\right)$, which only needs an $E$
instruction. 

Next, the asymptotic behavior of the length of the canonical string
is studied. Let $\rho$ be the probability that an element of the
adjacency matrix is 1, i.e., it is an active cell. It will be assumed
that each cell is independent of all the others. The behavior for
large sparse graphs is studied, i.e. $N\to\infty,\;\rho\to0$. The
average Manhattan distance $\delta$ from an active cell to the nearest
active cell can be approximated by (see Appendix A):

\begin{equation}
\mathbb{E}\left[\delta\right]\sim\frac{\sqrt{\pi}}{2\sqrt{2}}\,\rho^{-1/2}\quad\text{as }N\to\infty,\;\rho\to0
\end{equation}

The average number of instructions required to process an active cell
may be approximated by $\mathbb{E}\left[\delta\right]+1$ because
$\delta$ pointer movement instructions must be executed, plus an
$E$ instruction. It must be highlighted that the actual number of
instructions is higher because the closest other active cell may have
been already processed by the algorithm. The average number of active
cells in the adjacency matrix is $N^{2}\rho$. Therefore the length
of the canonical string can be approximated by the product of both
quantities:

\begin{equation}
\mathbb{E}\left[\left|I_{G}\right|\right]\sim\frac{\sqrt{\pi}}{2\sqrt{2}}\,N^{2}\sqrt{\rho}\quad\text{as }N\to\infty,\;\rho\to0\label{eq:ApproxCanonicalStringLength}
\end{equation}
where $\left|\cdot\right|$ stands for the length of a string, and
the term corresponding to the $E$ instruction has been dropped because
it becomes negligible in the limit. Equation (\ref{eq:ApproxCanonicalStringLength})
shows that for large, sparse graphs, the length of the canonical string
$I_{G}$ is much smaller than the length of the binary string representation
which is $N^{2}$. Consequently, the proposed approach substantially
compresses the structural information contained in the adjacency matrix,
while it remains a reversible transformation.

\subsection{Topological properties}

The proposed representation ensures that similar adjacency matrices
are mapped to similar strings. This is a fundamental advantage for
processing with deep learning models. Given an adjacency matrix $M$,
an arbitrary cell $\left(i,j\right)$, and an instruction string $w$
that represents $M$, let us consider the two cases that may arise
when flipping the content of the cell:
\begin{itemize}
\item If $M\left(i,j\right)=0$, then the path formed by the cells that
the pointer visits as $w$ is executed may be considered. Let $\left(i',j'\right)$
be the cell within the path which is closest in Manhattan distance
terms to $\left(i,j\right)$. Let $\delta$ be such smallest Manhattan
distance. Then a substring may be inserted in $w$ at the point where
$\left(i',j'\right)$ is visited. The substring moves the pointer
from $\left(i',j'\right)$ to $\left(i,j\right)$, followed by an
$E$ instruction, and finally goes back to $\left(i',j'\right)$.
The inserted substring adds $2\delta+1$ instructions to the string
length, and sets the element at $\left(i,j\right)$ to 1, while leaving
all the other elements with their previous values.
\item If $M\left(i,j\right)=1$, then there must be an $E$ instruction
in $w$ that sets that element to 1. Let $\left(i_{prev},j_{prev}\right)$
be the cell which is set to 1 by the previous $E$ instruction, and
$\left(i_{next},j_{next}\right)$ be the cell which is set to 1 by
the next $E$ instruction. Consequently, the substring of $w$ from
the previous $E$ instruction to the next $E$ instruction may be
substituted by a substring of movement instructions that take the
pointer from $\left(i_{prev},j_{prev}\right)$ to $\left(i_{next},j_{next}\right)$
by the shortest path according to Manhattan distance. From the triangle
inequality that Manhattan distance fulfills, it can be inferred that
the new substring is no longer than the old one. Therefore, the overall
string length cannot increase.
\end{itemize}
Together, the two above cases show that a minimal change in the adjacency
matrix corresponds to a small modification in the representing string,
as measured by the Levenshtein distance between strings. In other words,
a local change in the graph structure is associated with a local change
in the string. Therefore, local patterns in the graph are translated
into substrings within the representing string.

\section{Computational experiments\label{sec:Computational-experiments}}

In this section the results of some tentative experiments are reported. Subsection \ref{subsec:Dataset-description} explains how the synthetic dataset has been generated. Subsection \ref{subsec:Description-models} describes the neural transformer architectures that have been tested. After that, Subsection \ref{subsec:Experimental-setup} details the experimental setup. Finally, Subsection \ref{subsec:Results} discusses the obtained results.

\begin{center}
\begin{figure}
\begin{centering}

\includegraphics[width=8.5cm]{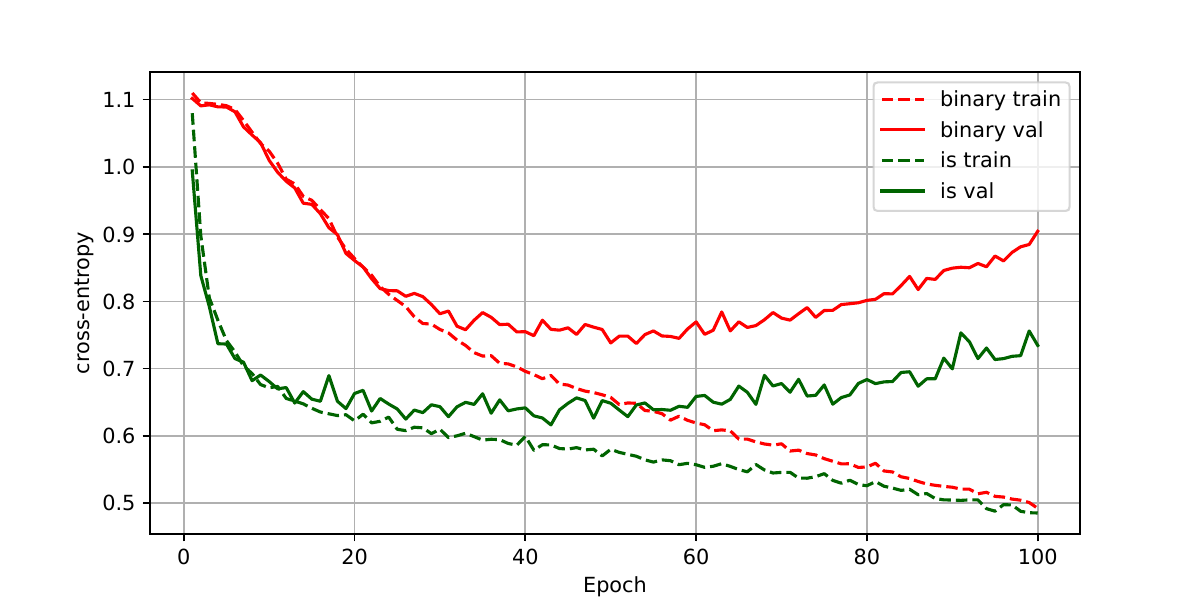}\includegraphics[width=8.5cm]{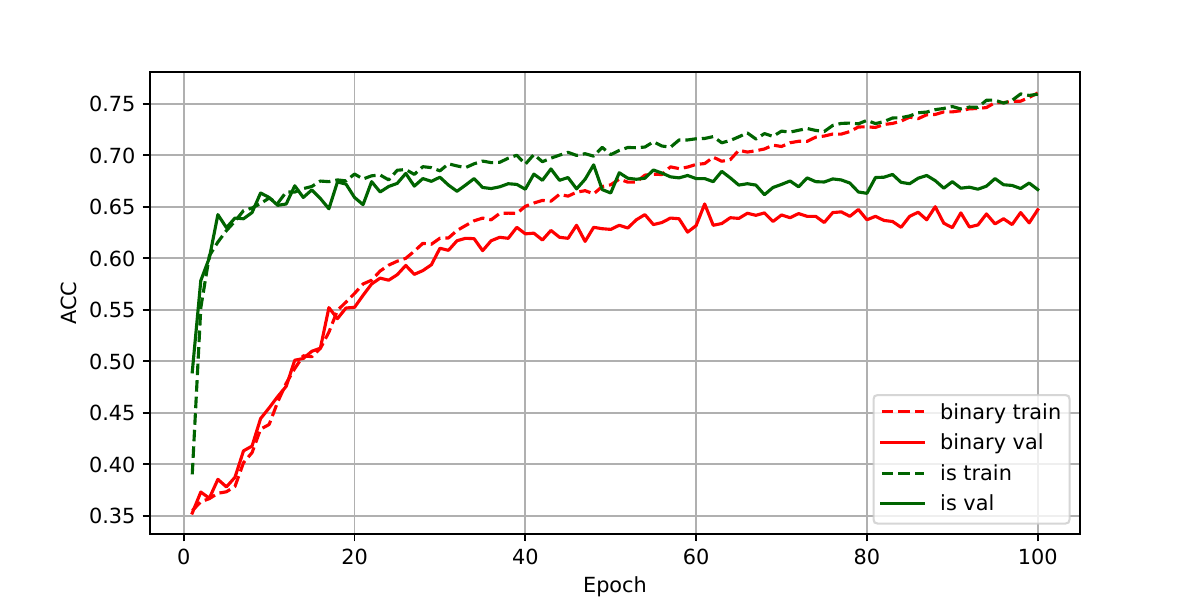}

\includegraphics[width=8.5cm]{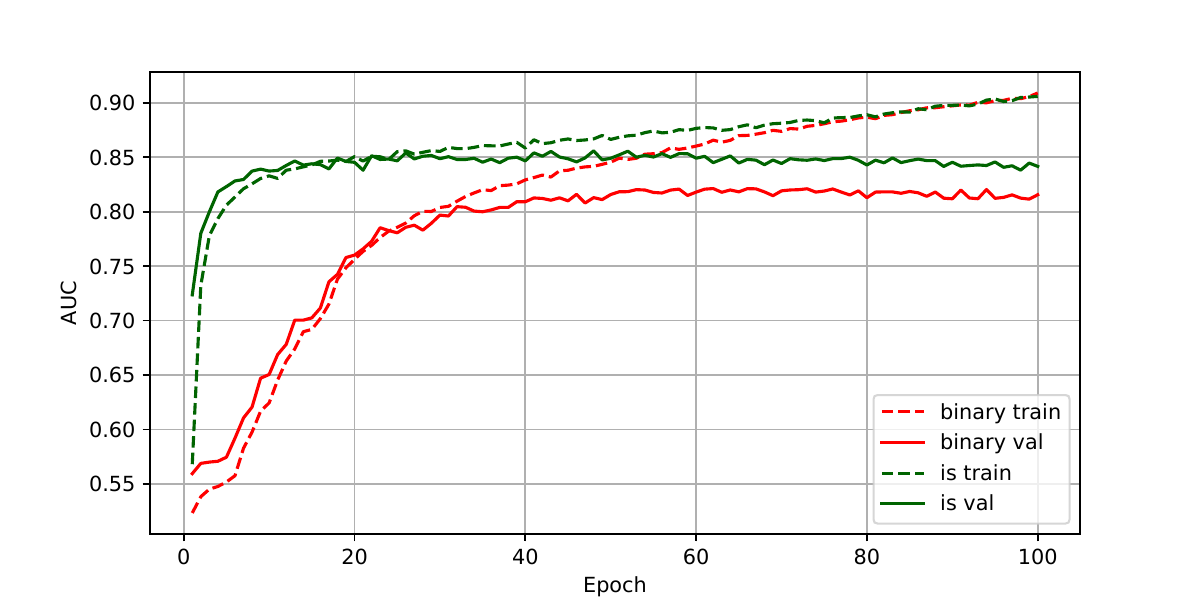}\includegraphics[width=7cm]{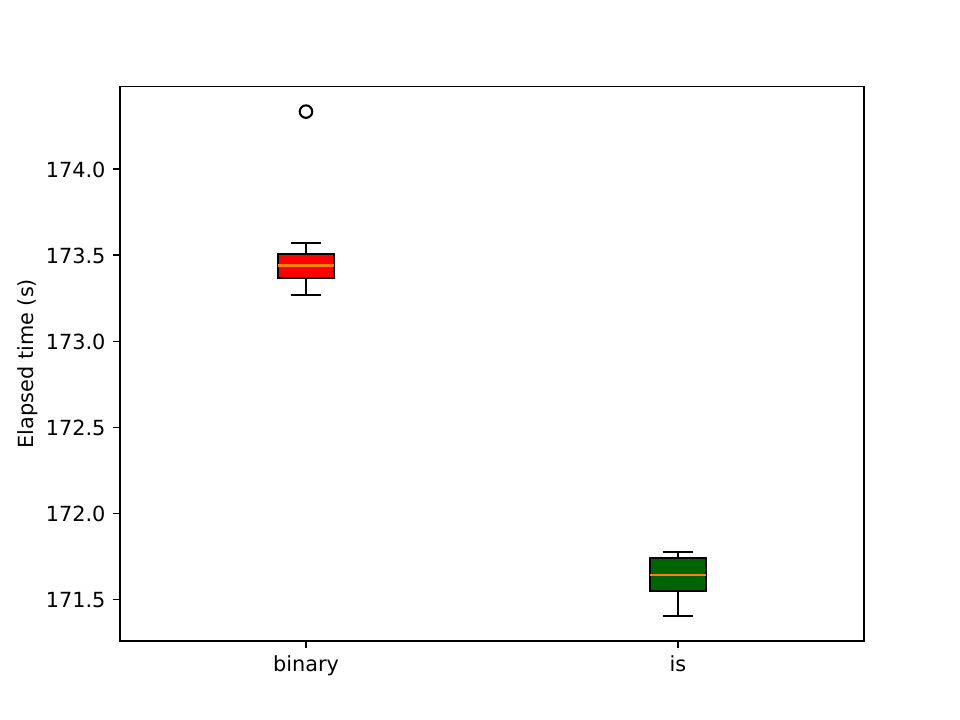}

\par\end{centering}
\caption{\label{fig:Results-4-2-128}Results for the models with $H=4$, $M=2$,
$F=128$.}
\end{figure}
\par\end{center}

\begin{center}
\begin{figure}
\begin{centering}

\includegraphics[width=8.5cm]{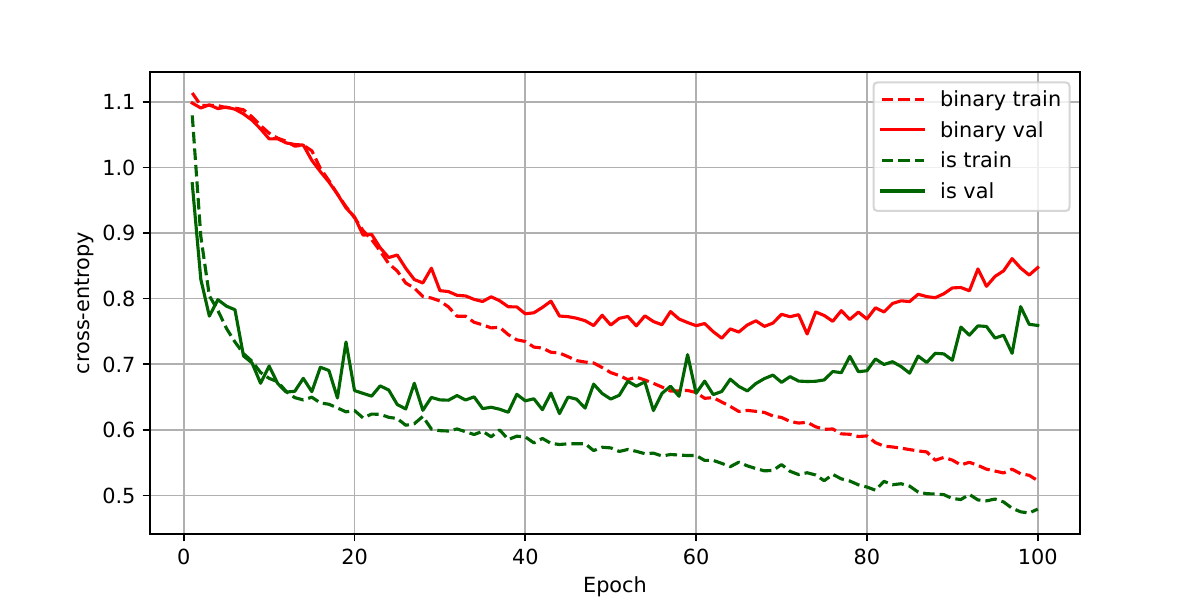}\includegraphics[width=8.5cm]{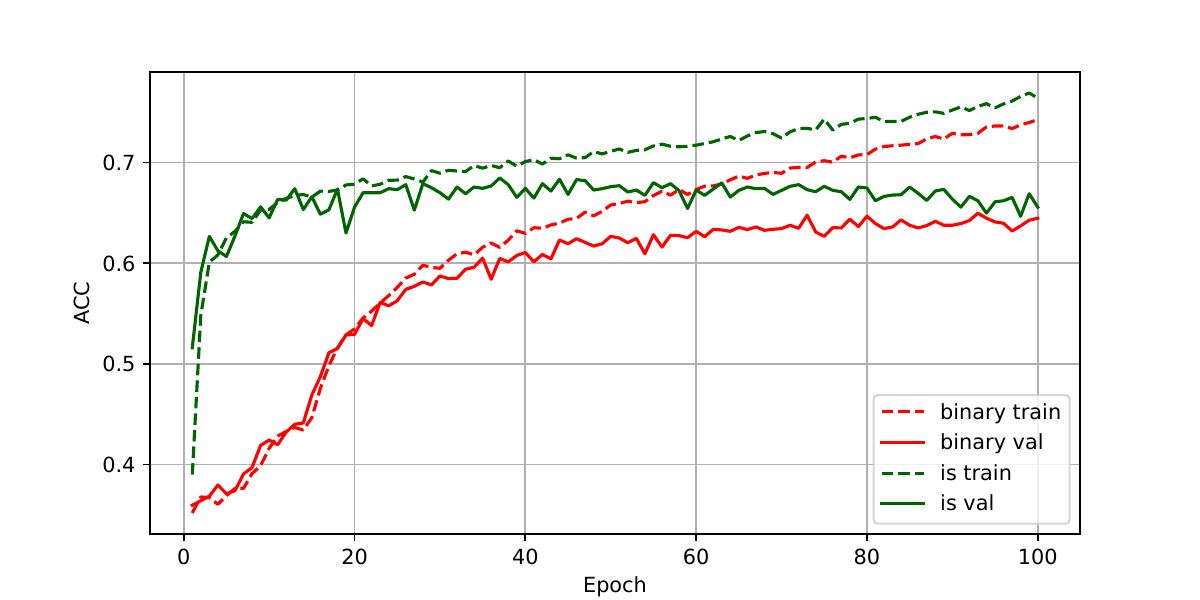}

\includegraphics[width=8.5cm]{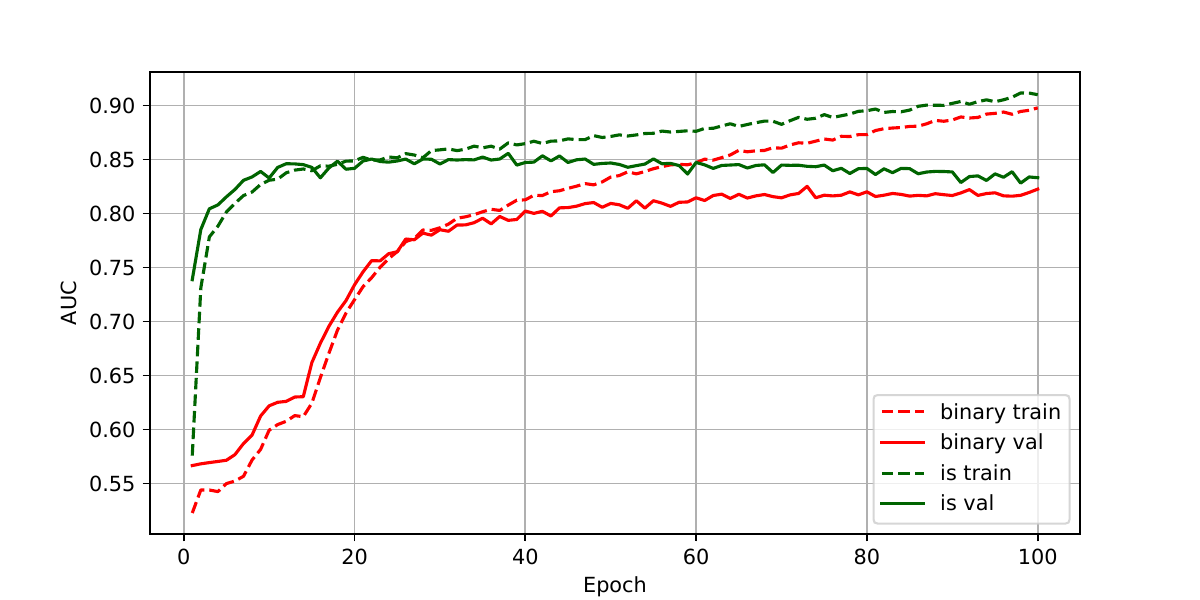}\includegraphics[width=7cm]{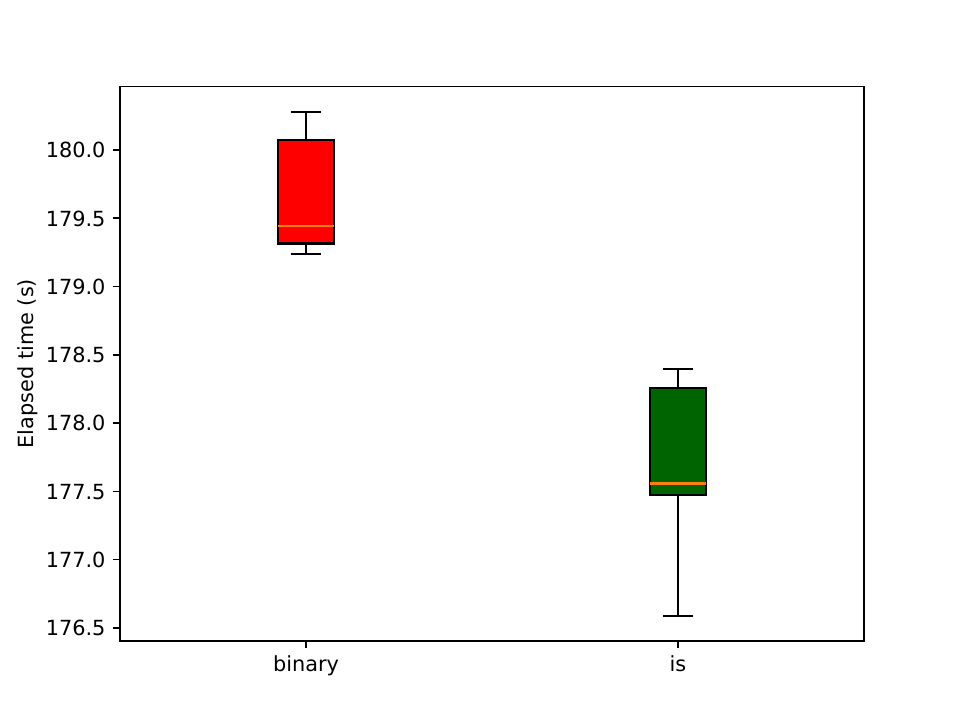}

\par\end{centering}
\caption{\label{fig:Results-4-2-256}Results for the models with $H=4$, $M=2$,
$F=256$.}
\end{figure}
\par\end{center}

\begin{center}
\begin{figure}
\begin{centering}

\includegraphics[width=8.5cm]{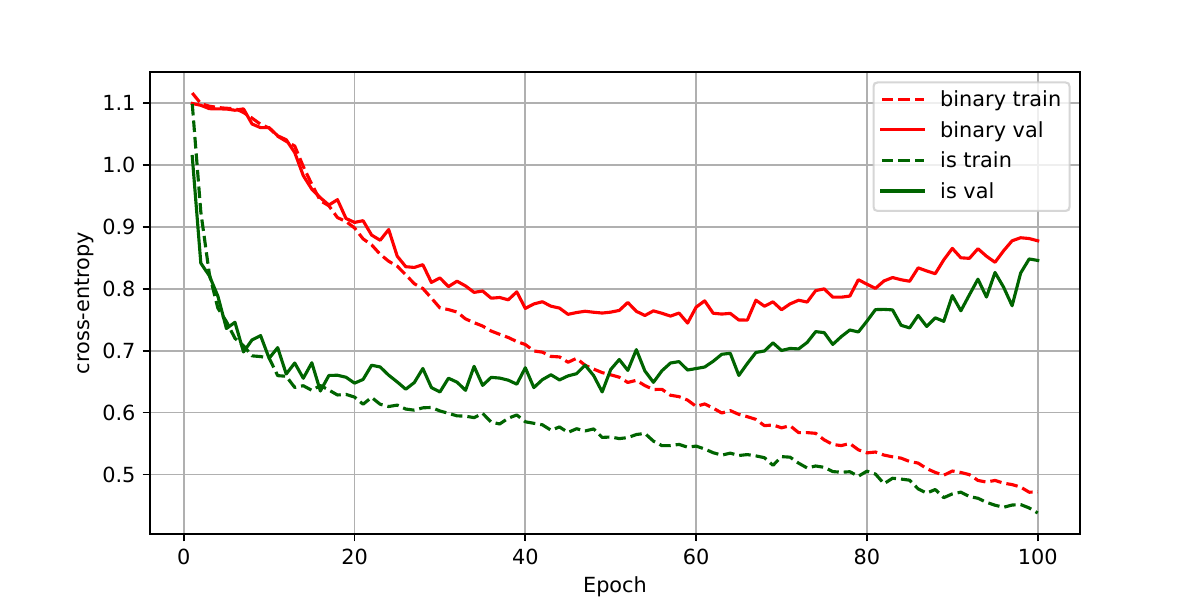}\includegraphics[width=8.5cm]{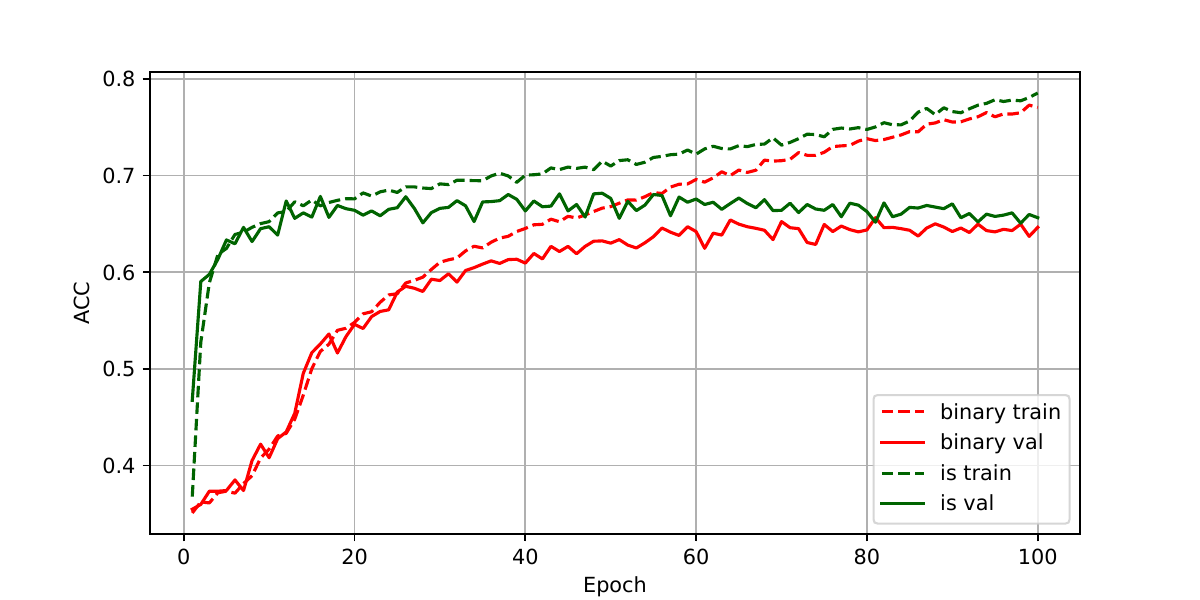}

\includegraphics[width=8.5cm]{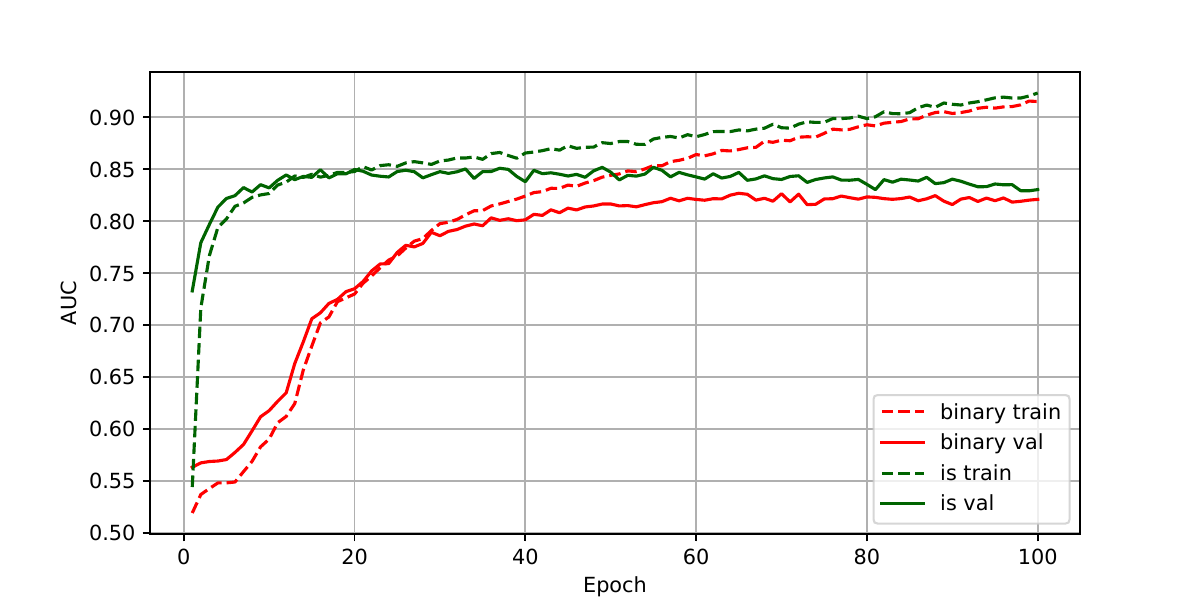}\includegraphics[width=7cm]{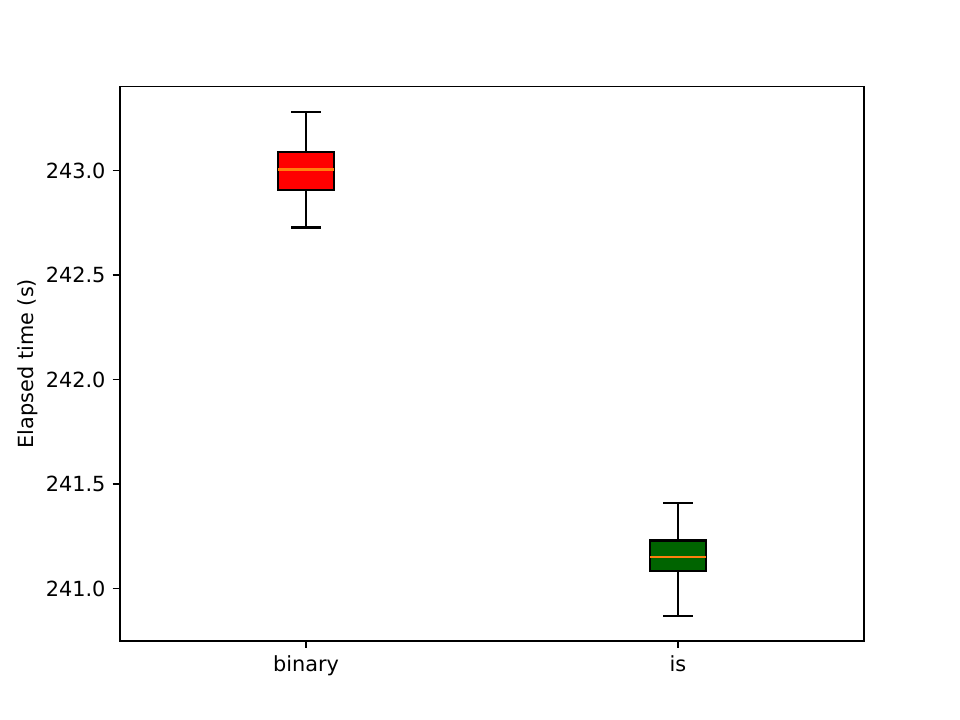}

\par\end{centering}
\caption{\label{fig:Results-4-3-128}Results for the models with $H=4$, $M=3$,
$F=128$.}
\end{figure}
\par\end{center}

\begin{center}
\begin{figure}
\begin{centering}

\includegraphics[width=8.5cm]{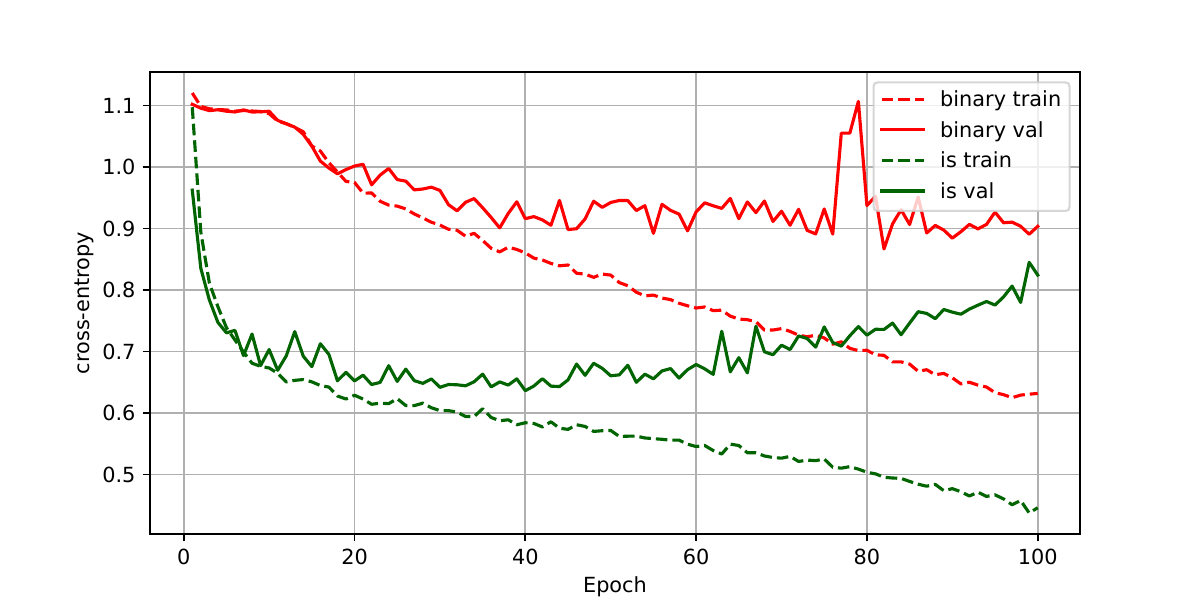}\includegraphics[width=8.5cm]{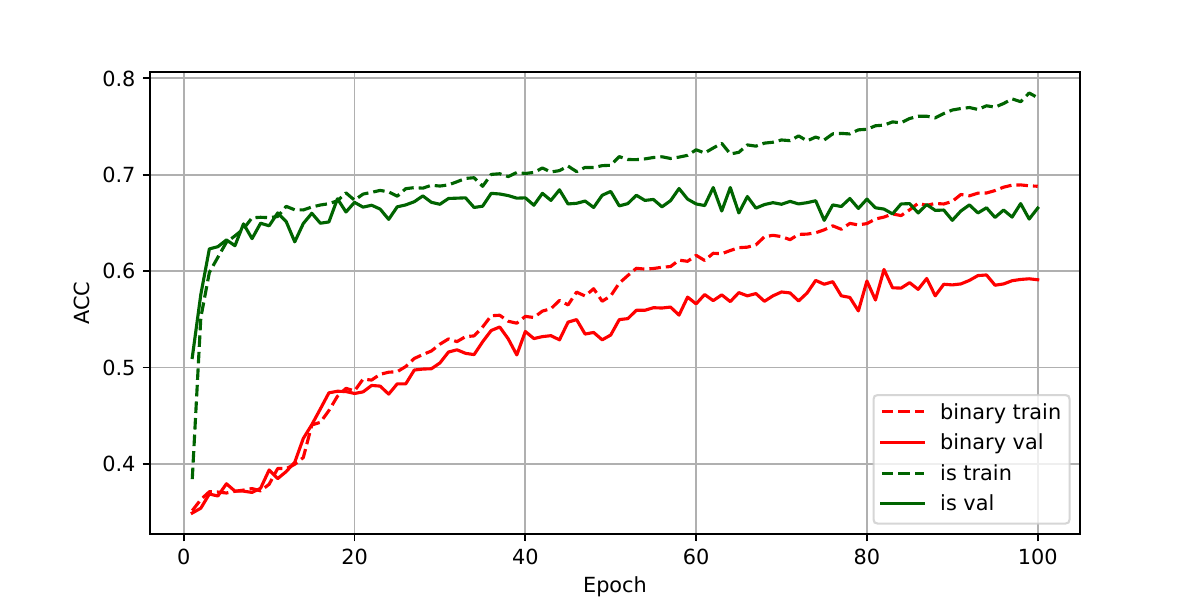}

\includegraphics[width=8.5cm]{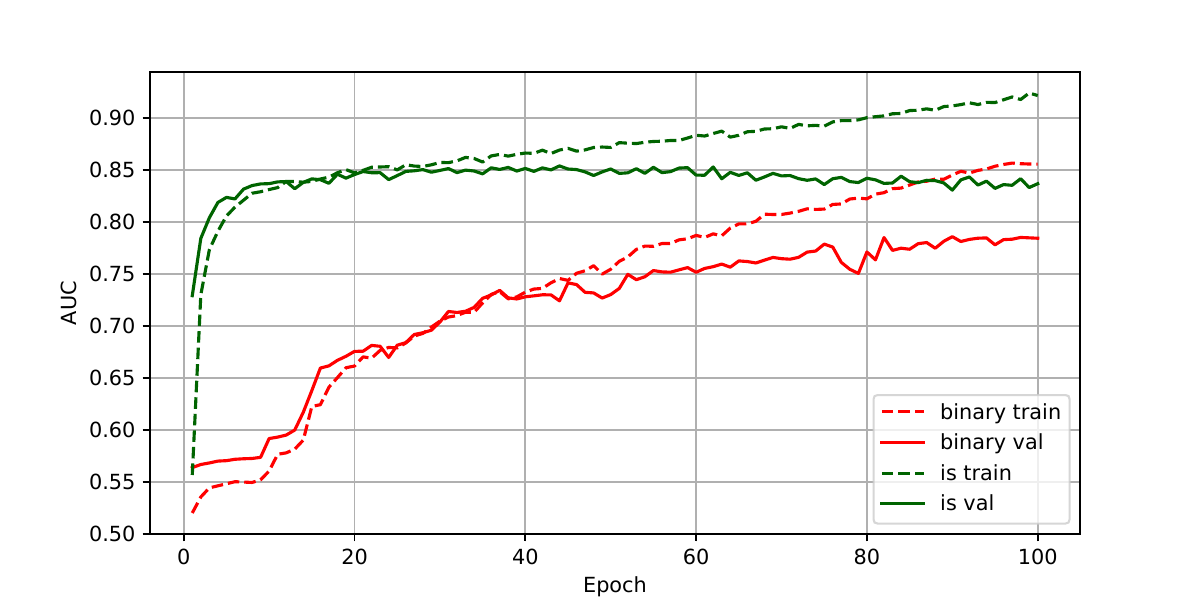}\includegraphics[width=7cm]{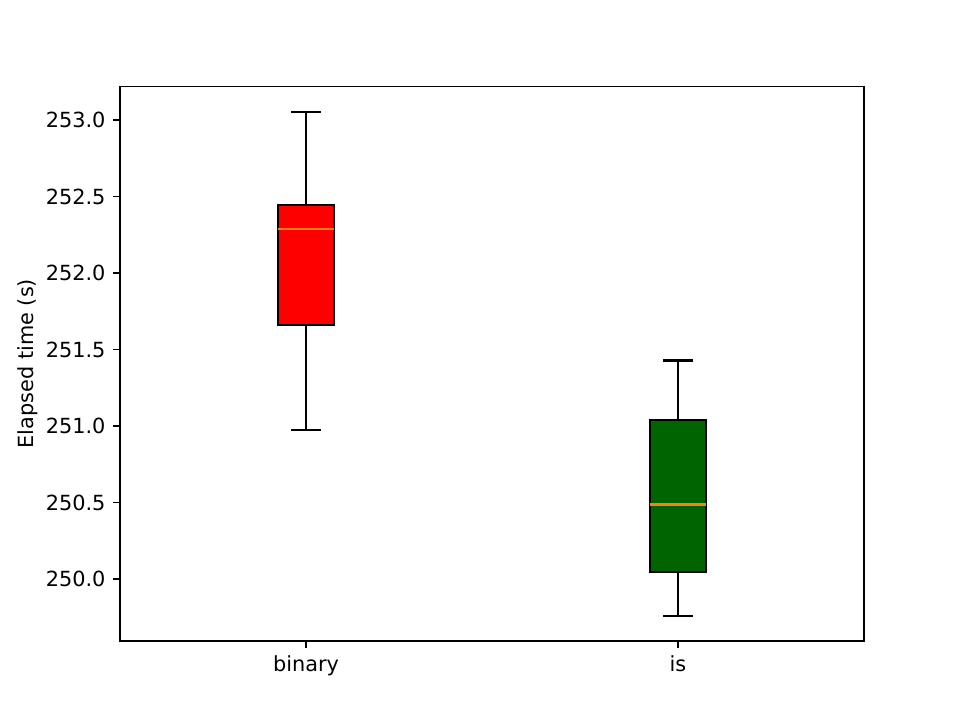}

\par\end{centering}
\caption{\label{fig:Results-4-3-256}Results for the models with $H=4$, $M=3$,
$F=256$.}
\end{figure}
\par\end{center}

\begin{center}
\begin{figure}
\begin{centering}

\includegraphics[width=8.5cm]{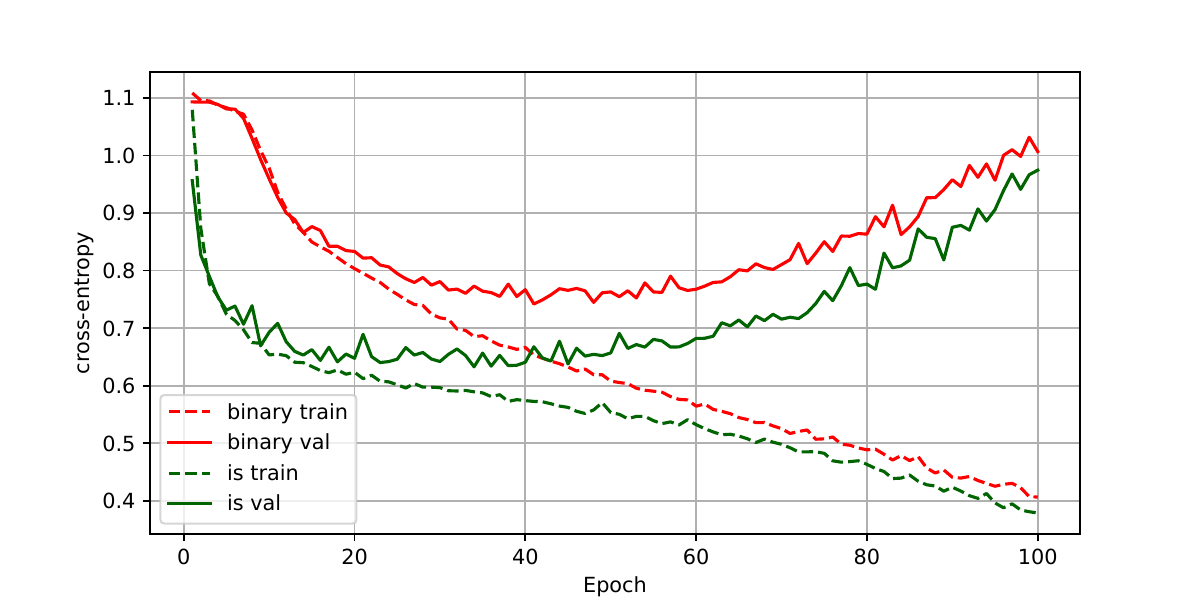}\includegraphics[width=8.5cm]{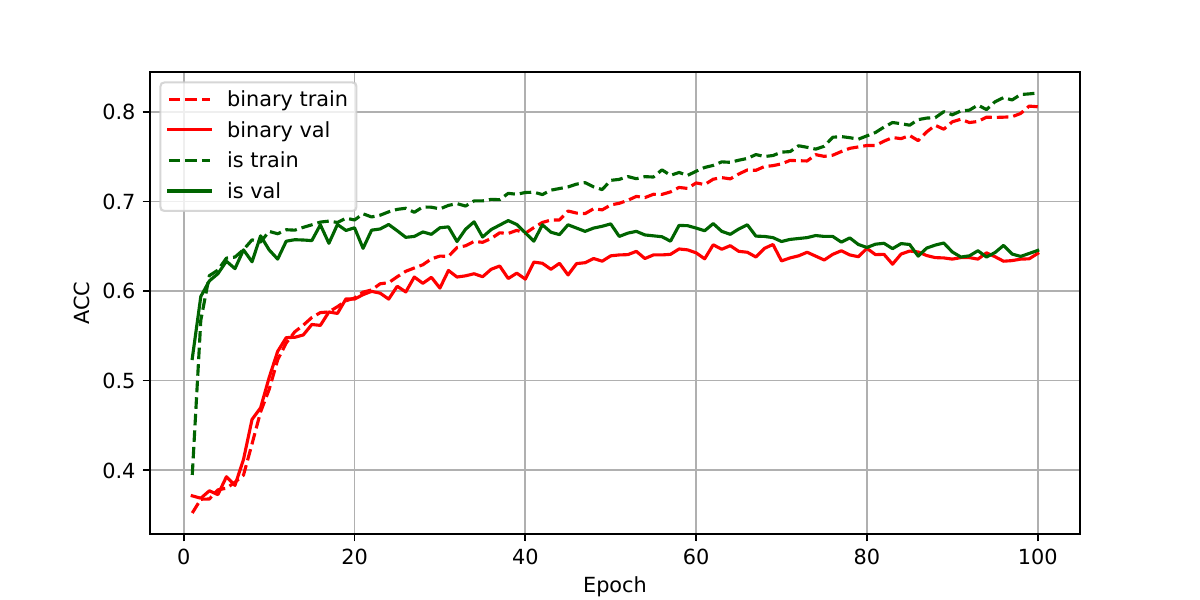}

\includegraphics[width=8.5cm]{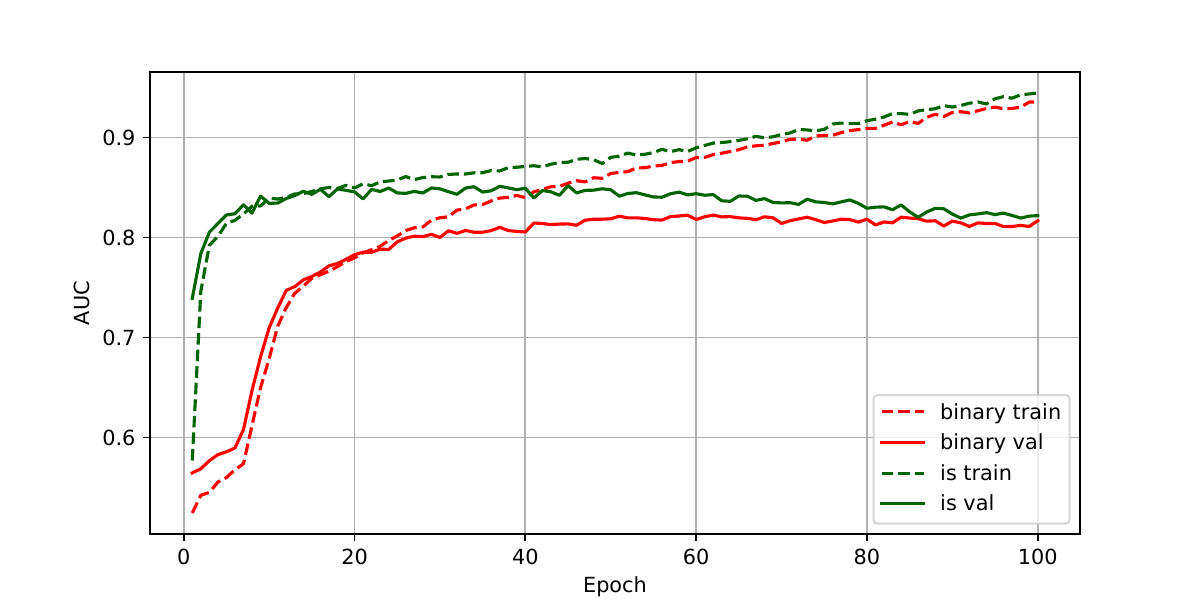}\includegraphics[width=7cm]{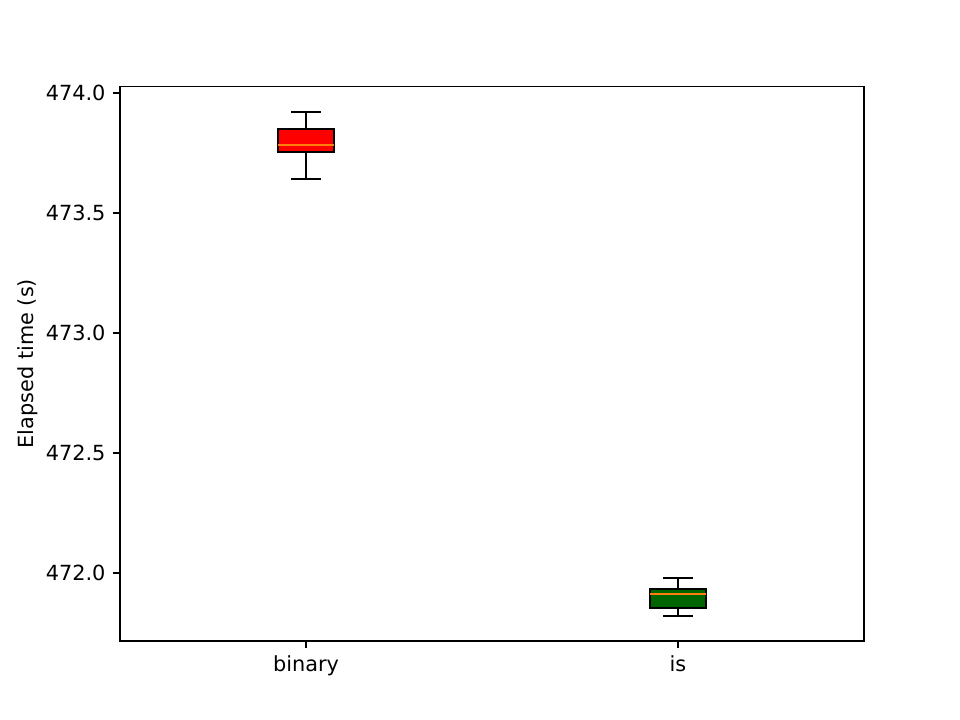}

\par\end{centering}
\caption{\label{fig:Results-16-2-128}Results for the models with $H=16$, $M=2$,
$F=128$.}
\end{figure}
\par\end{center}

\begin{center}
\begin{figure}
\begin{centering}

\includegraphics[width=8.5cm]{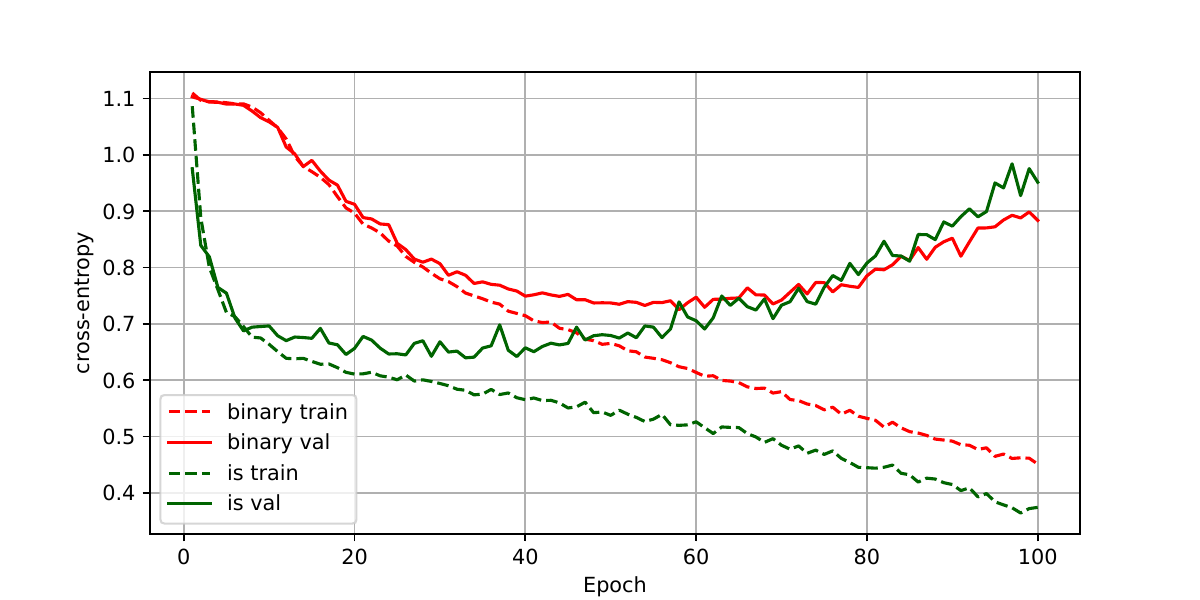}\includegraphics[width=8.5cm]{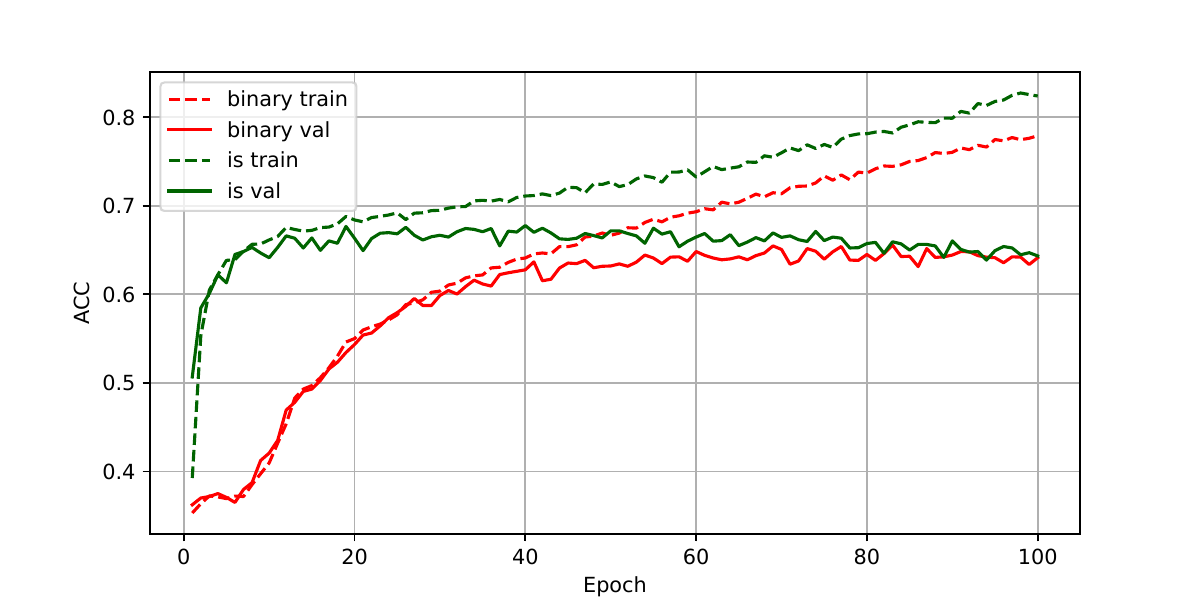}

\includegraphics[width=8.5cm]{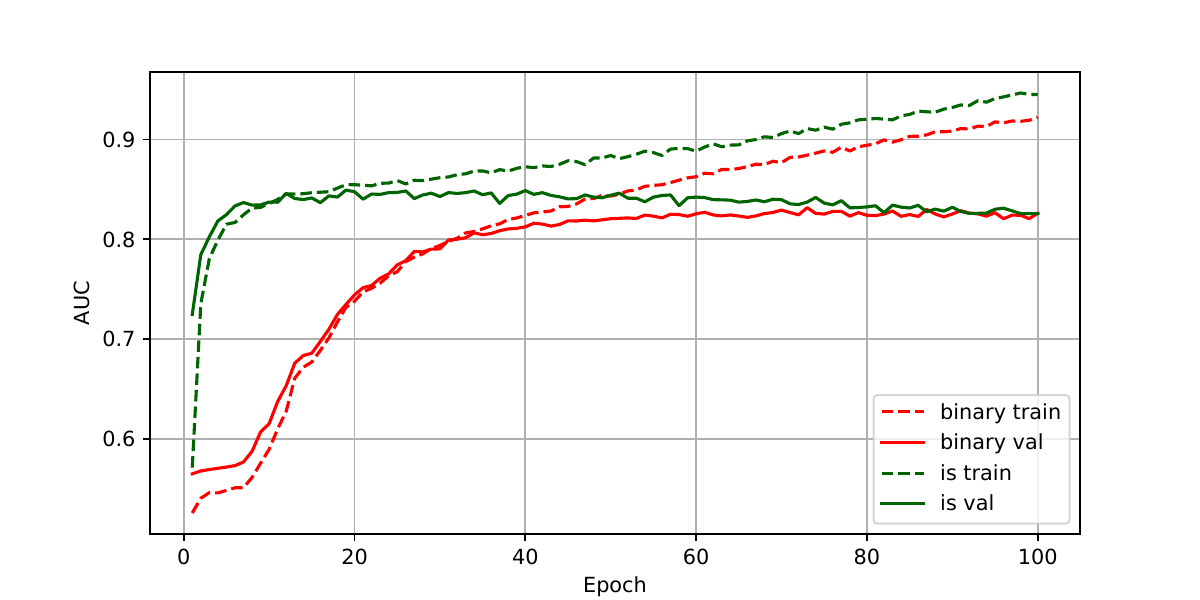}\includegraphics[width=7cm]{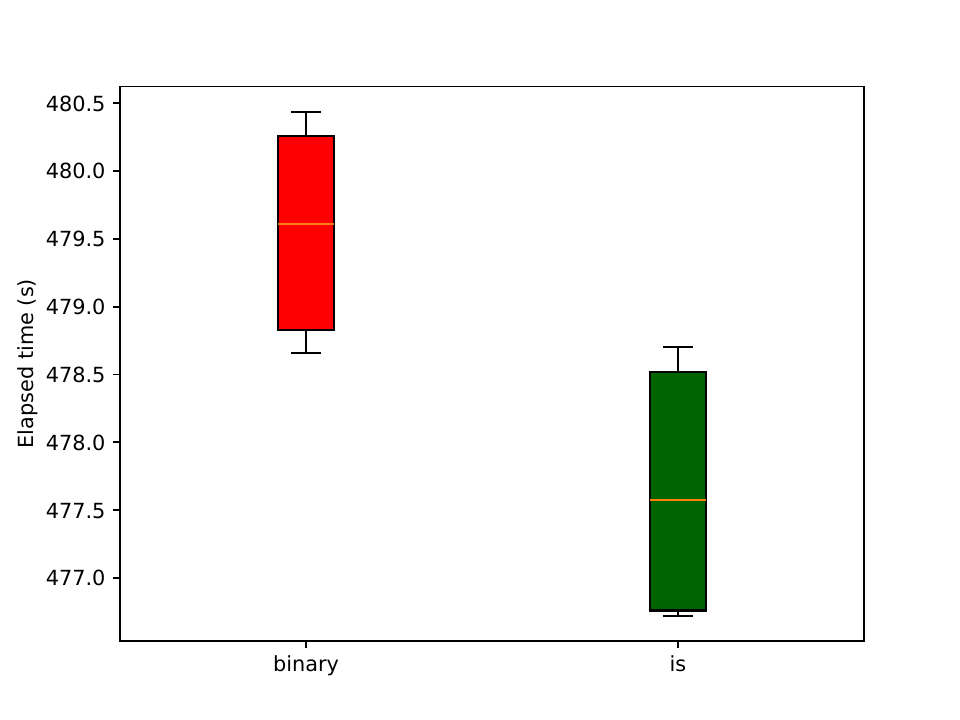}

\par\end{centering}
\caption{\label{fig:Results-16-2-256}Results for the models with $H=16$, $M=2$,
$F=256$.}
\end{figure}
\par\end{center}

\begin{center}
\begin{figure}
\begin{centering}

\includegraphics[width=8.5cm]{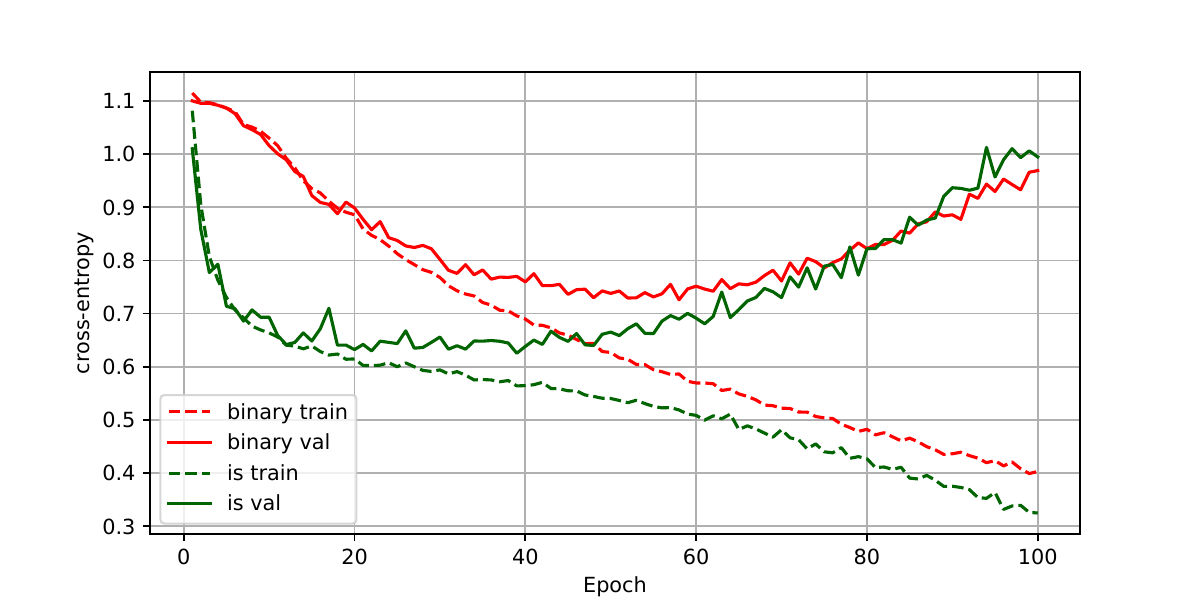}\includegraphics[width=8.5cm]{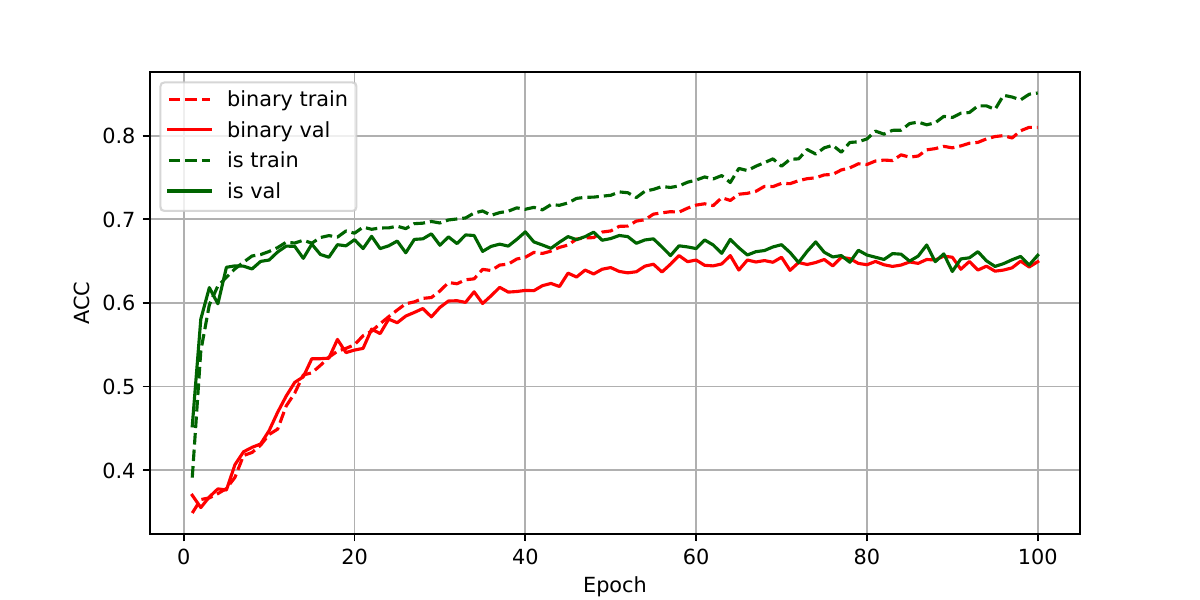}

\includegraphics[width=8.5cm]{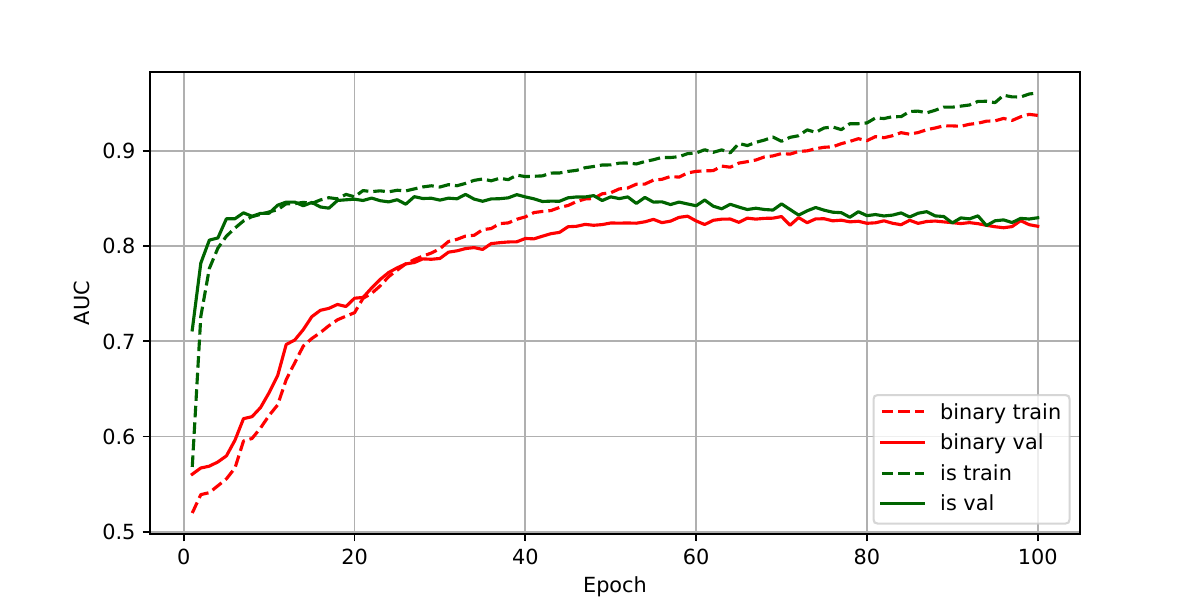}\includegraphics[width=7cm]{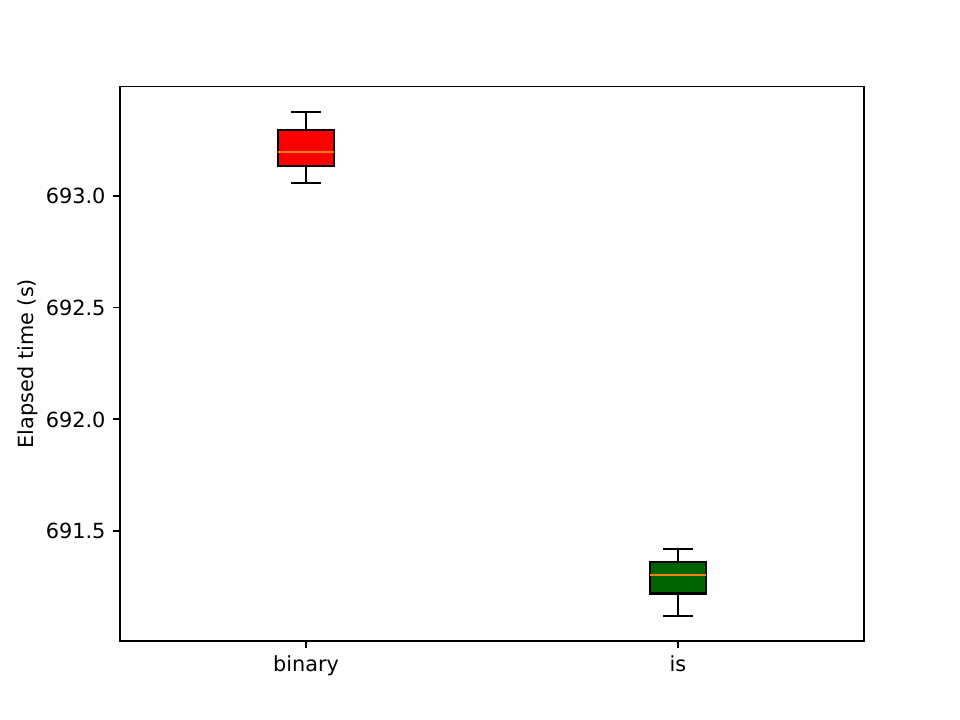}

\par\end{centering}
\caption{\label{fig:Results-16-3-128}Results for the models with $H=16$, $M=3$,
$F=128$.}
\end{figure}
\par\end{center}

\begin{center}
\begin{figure}
\begin{centering}

\includegraphics[width=8.5cm]{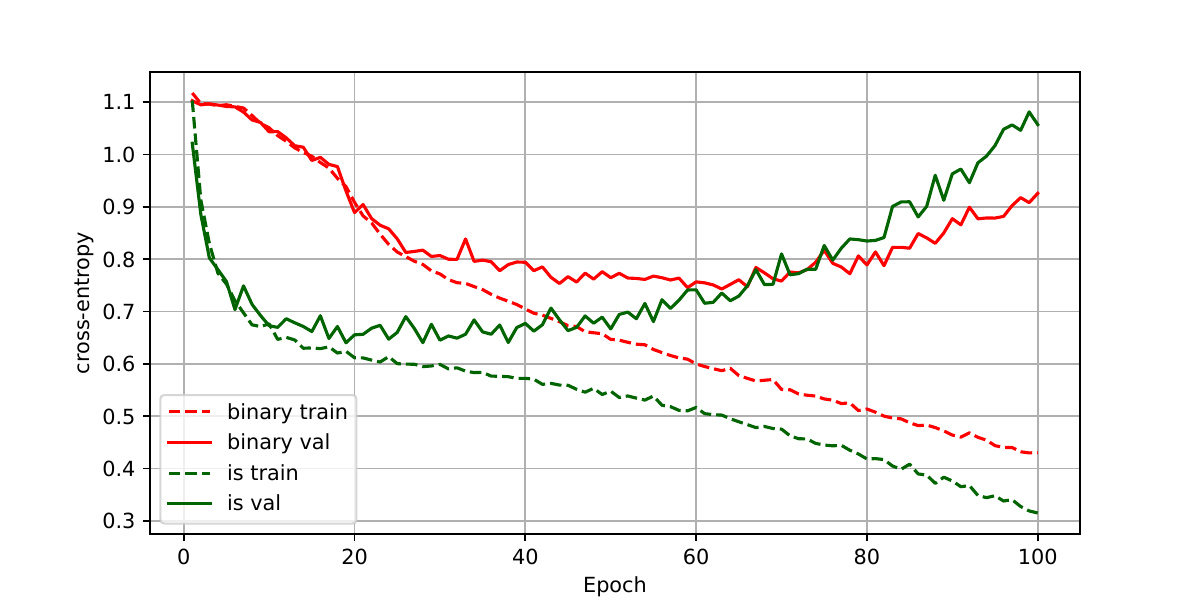}\includegraphics[width=8.5cm]{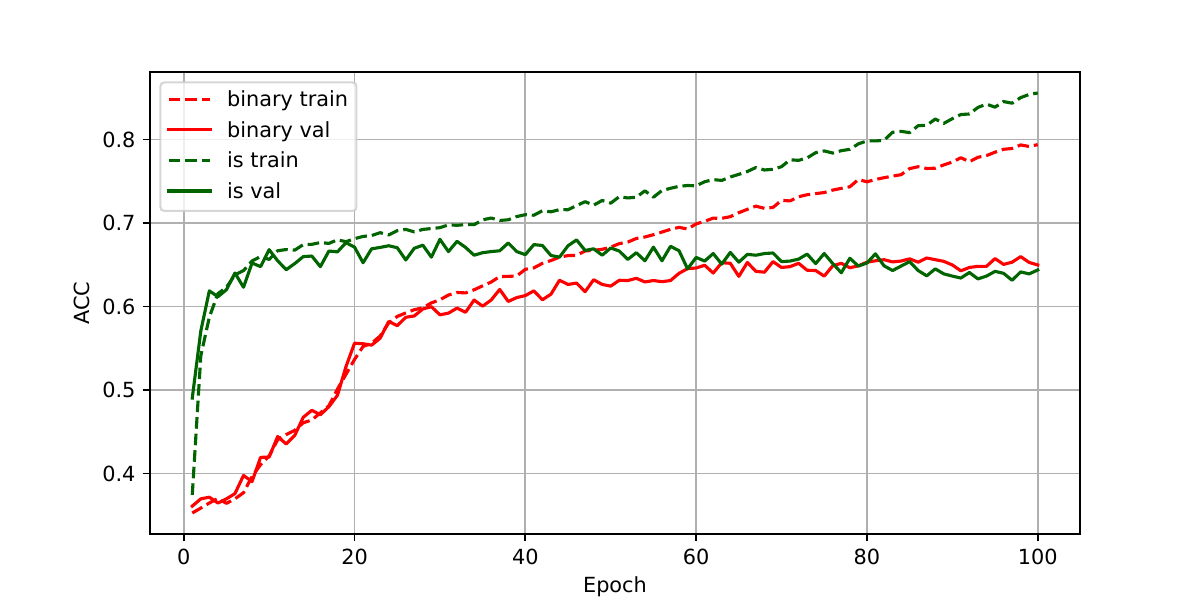}

\includegraphics[width=8.5cm]{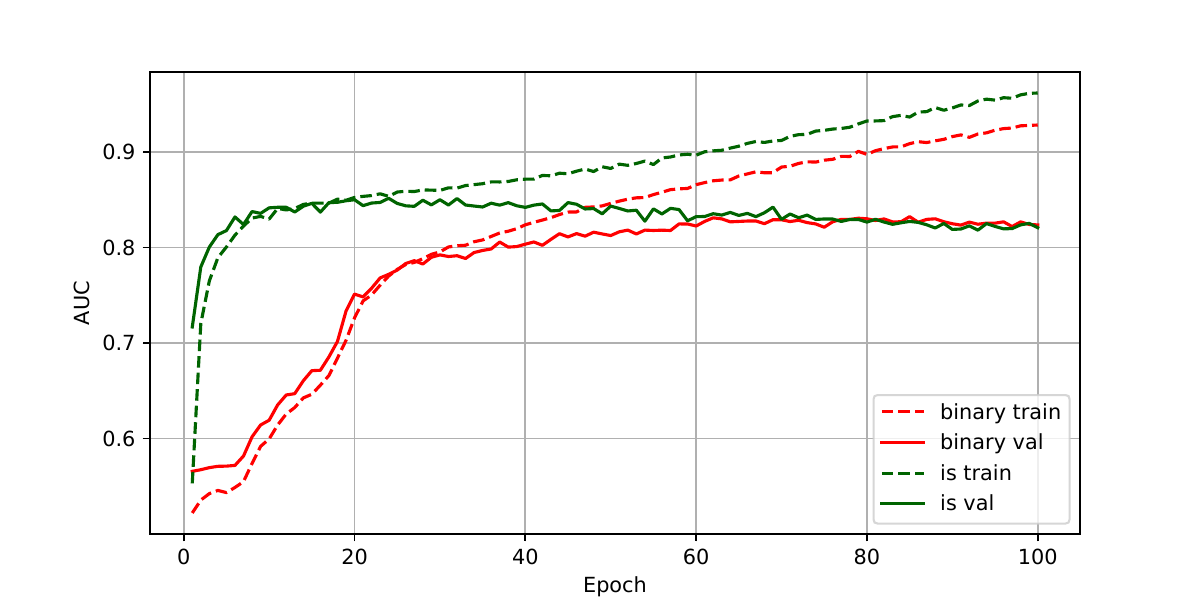}\includegraphics[width=7cm]{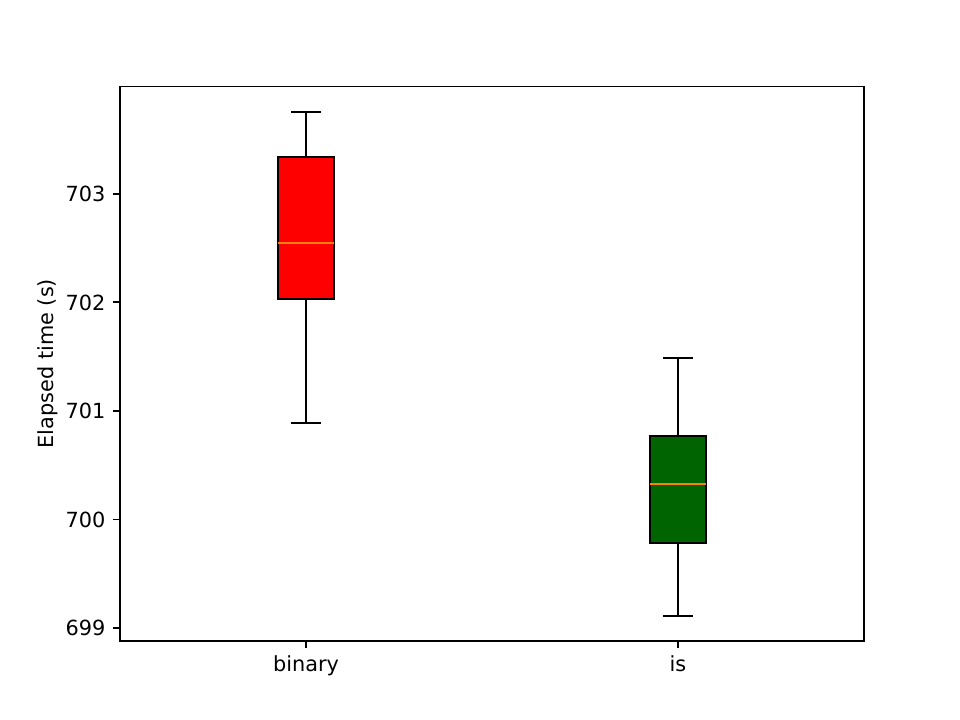}

\par\end{centering}
\caption{\label{fig:Results-16-3-256}Results for the models with $H=16$, $M=3$,
$F=256$.}
\end{figure}
\par\end{center}

\subsection{Dataset description\label{subsec:Dataset-description}}

The synthetic dataset that has been tailored to test our approach
is detailed next. A classification task has been designed, where three
classes of undirected graphs must be recognized. Each graph represents
adjacency relations in the three-dimensional space. A sequence of
points in 3D space is defined for each graph. Each point is associated
with a vertex of the graph. Then, an Euclidean distance threshold
$\zeta$ is established. Finally, the adjacency matrix of the graph
contains an edge between each pair of vertices associated with points
that are closer than the threshold $\zeta$ in 3D space. Each class
of graphs corresponds to a different way to generate a random sequence
of points in 3D:
\begin{enumerate}
\item A single random walk whose length is $2+\psi$, where $\psi\sim\textrm{Poisson}\left(10\right)$.
Each step in the random walk is defined by a 3D displacement vector
$\mathbf{v}\in\mathbb{R}^{3}$ from the previous point, where $\mathbf{v}\sim\textrm{Gauss}\left(\mathbf{0},\mathbf{I}\right)$. 
\item Two random walks, each with possibly different lengths $1+\psi_{1}$,
$1+\psi_{2}$, where $\psi_{1}\sim\textrm{Poisson}\left(5\right)$,
$\psi_{2}\sim\textrm{Poisson}\left(5\right)$. A step with a large
displacement $\mathbf{v}\sim\textrm{Gauss}\left(\mathbf{0},10\cdot\mathbf{I}\right)$
is taken between the two random walks. 
\item A torus is considered with the following parametric equations:\\
\begin{equation}
x\left(\theta,\phi\right)=\left(K+k\cos\theta\right)\cos\phi
\end{equation}
\\
\begin{equation}
y\left(\theta,\phi\right)=\left(K+k\cos\theta\right)\sin\phi
\end{equation}
\\
\begin{equation}
z\left(\theta,\phi\right)=k\sin\theta
\end{equation}
\\
where $K=10$ is the major radius and $k=1$ is the minor radius.
The length of the sequence is $2+\psi$, where $\psi\sim\textrm{Poisson}\left(10\right)$.
For the $i$-th point in the sequence, $i\in\left\{ 0,2,....,1+\psi\right\} $,
the first parameter $\theta$ is set to $\frac{2\pi\left(\xi+i\right)}{2+\psi}$,
where $\xi\sim\textrm{Gauss\ensuremath{\left(0,1\right)}}$. This
makes $\theta$ span its entire range $\left[0,2\pi\right]$ with
added Gaussian noise. The second parameter is set to $\phi\sim\textrm{Uniform}\left(0,2\pi\right)$,
which is a uniform random draw from its entire range of values.
\end{enumerate}
For each class, the distance threshold $\zeta$ is set to the 20th
percentile of the distribution of distances, so that approximately
20\% of the elements of each adjacency matrix are set to 1. This way,
each class contains graphs that comprise different structures.

\subsection{Description of the models\label{subsec:Description-models}}

A range of simple neural transformer models has been selected in order
to evaluate the advantages of the proposed graph representation methodology
by instruction strings (\emph{is}), as compared to the standard binary
string representation (\emph{binary}). Each transformer model processes
an input string of characters and outputs three class scores associated
with the three classes in the dataset. The architecture of the transformers
comprises:
\begin{enumerate}
\item An input embedding layer.
\item A positional encoding layer.
\item A transformer encoder with $M$ layers. Each layer has $H$ heads,
and a feedforward network model dimension $F$. The dropout is set to
$0.1$.
\item An output linear layer.
\end{enumerate}
The embedding dimension is 64 for all layers except the linear layer.
Eight model sizes have been tested, corresponding to $M\in\left\{ 2,3\right\} $,
$H\in\left\{ 4,16\right\} $, $F\in\left\{ 128,256\right\} $. 

\subsection{Experimental setup\label{subsec:Experimental-setup}}

A dataset of 3000 samples with 1000 samples per class has been drawn
according to Subsection \ref{subsec:Dataset-description}. The graphs
of the dataset have been represented with binary strings (baseline)
and instruction strings (our proposal). For each string representation
and model size, a 10-fold cross-validation has been run. Four NVidia
A100 GPUs with 40GB VRAM each have been employed for the experiments.
The cross-entropy has been employed as the training loss. The Adam
optimizer with learning rate $0.001$ and batch size 64 has been chosen,
with 100 training epochs. 

Four performance metrics have been measured:
\begin{itemize}
\item The cross-entropy loss.
\item The classification accuracy ($ACC$).
\item The Area Under the Curve ($AUC$) following the One Versus Rest criterion.
\item The overall computation time, including training and validation.
\end{itemize}
For each performance metric, the values for the 10 cross-validation
folds have been collected. The means over the 10 folds on the training
and validation sets have been computed for the first three metrics.

\subsection{Results\label{subsec:Results}}

In this subsection the results of the experiments are reported. Figures
\ref{fig:Results-4-2-128} to \ref{fig:Results-16-3-256} depict the
obtained results. The cross-entropy loss results indicate that the
networks learn correctly for several tens of epochs. After that, the
overfitting regime starts, in all cases well before the 100 epoch
limit, marked by an increase in the validation set cross-entropy.
Therefore, the overall behavior of the learning process is as expected.
The models that use the instruction string representation tend to
learn faster while they attain a better validation set cross-entropy,
which demonstrates that learning is easier with our proposal.

These results are further confirmed by the downstream task metrics,
namely the accuracy and the Area Under the Curve. Both metrics show
that the instruction string representation consistently obtains the
best classification performance. The models that run on the binary
representation cannot reach the performance values of our proposal,
no matter how long they are trained.

Regarding the computation time, our proposal is the fastest for all
model sizes. The differences are not very significant, which may be
due to the fact that the strings are not very long. More acute differences
may be encountered for longer strings corresponding to larger graphs.

\section{Conclusion\label{sec:Conclusion}}

A new method to represent the structure of a graph by a sequence of
instructions has been proposed. Each string in the regular language
formed by all strings over an alphabet of five instructions represents
a graph. Conversely, any graph can be represented by an infinite set
of strings of such language. A canonical string is defined for each
graph. Consequently, a reversible transformation between strings and
graphs is defined. The representation is compact because a reduced
number of instructions is needed to represent a graph, in particular
for large, sparse graphs. Moreover, small changes in a graph correspond
to small Levenshtein distances between the representing strings. Therefore,
local changes remain local as the transformation is applied.

Together, all these properties suggest that the proposed representation
is amenable for its use with deep learning models, in particular language
models. Tentative experiments indicate that this may well be the case.

\section*{Acknowledgment}
The author thankfully acknowledges the computer resources (Picasso
Supercomputer), technical expertise, and assistance provided by the
SCBI (Supercomputing and Bioinformatics) center of the University
of M\'alaga.

\section*{Appendix A}
For a large adjacency matrix with i.i.d. activations of density $ \rho\to 0$, the mean Manhattan distance $\delta$ from a typical active cell to its nearest other active cell scales like a constant times $\rho^{-1/2}$. More precisely, the asymptotic form is

\[
\mathbb{E}[\delta] \sim C\,\rho^{-1/2},\quad \rho\to 0,
\]
with \(C\) of order 1.

For a continuum approximation by a homogeneous Poisson point process of intensity \(\lambda = \rho\) on \(\mathbb{R}^2\), the number of other active cells inside the L1 ball of radius \(r\),
\[
B_1(0,r)=\{x\in\mathbb{R}^2:\|x\|_1\le r\},
\]
is Poisson with mean \(\lambda\,\text{Area}(B_1(0,r))= \lambda\cdot 2r^2\).

Hence the survival function of the nearest–neighbor L1 distance \(S\) for a typical point is
\[
\mathbb{P}(S>r)=\exp(-2\lambda r^2),\quad r\ge 0.
\]

Differentiating gives the probability density
\[
f_S(r)=4\lambda r\,e^{-2\lambda r^2},\quad r\ge 0.
\]

The mean nearest–neighbor distance in this model is
\[
\mathbb{E}[S]=\int_0^\infty r\,f_S(r)\,dr
=4\lambda\int_0^\infty r^2 e^{-2\lambda r^2}\,dr.
\]

Evaluating the integral yields
\[
\mathbb{E}[S]=\frac{\sqrt{\pi}}{2\sqrt{2}}\lambda^{-1/2} \approx 0.6267\,\lambda^{-1/2}.
\]

Identifying \(\lambda=\rho\) gives the asymptotics for the adjacency matrix model:
\[
\mathbb{E}[\delta] \sim \frac{\sqrt{\pi}}{2\sqrt{2}}\,\rho^{-1/2}\quad\text{as }\rho\to 0.
\]

\bibliographystyle{unsrtnat}
\bibliography{references}  






\end{document}